\documentclass[letterpaper, 10 pt, conference]{ieeeconf}  

\IEEEoverridecommandlockouts                              

\usepackage{amsmath}
\usepackage{amssymb}
\usepackage{xcolor}
\usepackage{graphicx}
\usepackage{comment}
\usepackage{subcaption}
\usepackage{caption}
\DeclareCaptionFont{8pt}{\fontsize{8pt}{12pt}\selectfont}
\newcommand{\name}{FlowMaps}
\newcommand{\todo}[1]{\textcolor{red}{#1}}
\definecolor{query}{HTML}{7EA6E0}
\definecolor{cond}{HTML}{AA0000}
\definecolor{cont}{HTML}{A680B8}

\title{\LARGE \bf
Dynamic Objects Relocalization in Changing Environments with Flow Matching}

\author{Francesco Argenziano$^{1,*}$, Miguel Saavedra-Ruiz$^{2,3,*}$, Sacha Morin$^{2,3}$, Daniele Nardi$^{1}$, and Liam Paull$^{2,3
}$
\thanks{*Authors contributed equally.}
\thanks{$^{1}$Department of Computer, Automation and Management
Engineering, Sapienza University of Rome, 00181 RM
Rome, Italy. {\tt\small \{argenziano, nardi\}@diag.uniroma1.it}}%
\thanks{$^{2}$Department of Computer Science and Operations Research, Université
de Montréal, Montréal, QC, Canada.}%
\thanks{$^{3}$Mila - Quebec AI Institute, Montréal, QC, Canada. {\tt\small \{miguel-angel.saavedra-ruiz, sacha.morin, paulll\}@mila.quebec}}
}

\begin{document}

\maketitle
\thispagestyle{empty}
\pagestyle{empty}

\begin{abstract}
Task and motion planning are long-standing challenges in robotics, especially when robots have to deal with dynamic environments exhibiting long-term dynamics, such as households or warehouses. In these environments, long-term dynamics mostly stem from human activities, since previously detected objects can be moved or removed from the scene. This adds the necessity to find such objects again before completing the designed task, increasing the risk of failure due to missed relocalizations. However, in these settings, the nature of such human-object interactions is often overlooked, despite being governed by common habits and repetitive patterns. Our conjecture is that these cues can be exploited to recover the most likely objects' positions in the scene, helping to address the problem of unknown relocalization in changing environments. To this end we propose \name, a 
model based on Flow Matching that is able to infer multimodal object locations over space and time. Our results present statistical evidence to support our hypotheses, opening the way to more complex applications of our approach. The code is publically available at \texttt{https://github.com/Fra-Tsuna/flowmaps}.
\end{abstract}

\section{INTRODUCTION}
Executing long-horizon tasks in real-world environments remains a major challenge for mobile robots. Even a simple instruction like ``fetch me the mug from the kitchen'' requires robust mapping, planning, and decision-making pipelines. Examples of such tasks in robotics
include \textbf{object navigation} and \textbf{object retrieval}: object navigation involves finding and navigating to a target object, whereas object retrieval also requires approaching, manipulating, and lastly delivering the object to a specified destination.\looseness-1

Recent advances in Large Language Models (LLMs) and Vision-Language Models (VLMs) have improved the performance of embodied agents in such tasks by enabling richer semantic queries. Specifically, LLMs support commonsense reasoning~\cite{huang2022language}, while VLMs provide compact, high-dimensional scene features (e.g., CLIP~\cite{clip}) that allow agents to better perceive and reason about objects in the environment~\cite{gu2024conceptgraphs}.\looseness-1

However, many approaches assume \emph{static} environments, where objects move only if the robot acts on them. This assumption is unrealistic: real-world scenes are \emph{dynamic}, as humans constantly move objects and reorganize rooms. Consequently, robots cannot rely on objects staying in their last observed locations, which poses a major risk for both navigation and retrieval tasks.

Humans, though, exhibit repetitive patterns~\cite{troje2008retrieving}: for example, a bottle taken from the kitchen table is likely placed on a desk and later returned to the kitchen sink. These \emph{semantically consistent patterns}~\cite{schmid2022panoptic, yugay2025gaussian} can be exploited to recover from missed retrievals by relocating objects at likely human-induced placements. Since these locations are inherently multimodal, a distributional approach is required.\looseness-1

We therefore propose \textbf{\name{}}, a Flow Matching~\cite{lipmanflow} model that recovers multimodal object distributions over space and time. Using a transformer-based map encoder and observations of object placements over time, \name{} infers plausible locations for query object classes, improving retrieval success.\looseness-1

We summarize our contributions as follows:
\begin{itemize}
\item \textbf{\name{}}: a Flow Matching–based model for multimodal object relocalization in long-term dynamic scenes.\looseness-1
\item \textbf{FlowSim}: a procedurally generated dataset that simulates multimodal object placements over time.
\item A qualitative evaluation empirically demonstrating the benefits of our approach, and a quantitative comparison against an MLP baseline.
\end{itemize}

This paper is organized as follows: Section~\ref{rw} reviews related work, Section~\ref{methodology} details \name{} and FlowSim, Section~\ref{results} presents the experimental results, and Section~\ref{conclusion} concludes.\looseness-1
\section{RELATED WORK}
\label{rw}
\textbf{Object navigation and object retrieval.} 
The goal of object navigation is to reach a target object specified by category or natural language, typically through semantic mapping and goal-conditioned exploration~\cite{chaplot2020object,sun}. Object retrieval extends this to finding, grasping, and delivering the object, with recent benchmarks enabling open-vocabulary mobile manipulation in realistic household environments~\cite{yenamadra2023homerobot,pmlr-v270-han25a}. Despite their effectiveness, many pipelines still assume static scenes or rely on short-horizon replanning when target objects move, which is a recurring source of failure in environments undergoing human activities.\looseness-1

\textbf{Flow Matching in robotics.}
Generative models are increasingly prominent in robotics, aided by large, cross-embodiment datasets such as Open-X Embodiment~\cite{o2024open}. Building on this, recent work has applied diffusion~\cite{chi2024diffusionpolicy, hou2024diffusion} and Flow Matching (FM)~\cite{chisari2025learning, zhang2024affordance} to learn action policies, while others have developed end-to-end vision-language-action systems~\cite{black2024pi0, kim2025openvla}. While these approaches have shown impressive capabilities, they largely focus on policy learning. In contrast, we leverage FM to recover spatio-temporal, multimodal posterior distributions over potential object placements rather than generate action tokens. To our knowledge, this is the first application of FM to posterior inference in this setting.
\section{METHODOLOGY} 
\label{methodology}
This section is organized as follows. Section~\ref{math} provides an overview
of Flow Matching, followed by Section~\ref{probform}, which outlines the problem formulation for dynamic objects relocalization. Section~\ref{data} details the data collection procedure, and Section~\ref{arch} gives a complete description of our system's architecture, with implementation details provided in Section~\ref{implementation}.

\vspace{-0.4cm}
\subsection{Preliminaries}
\label{math}

An \emph{Ordinary Differential Equation} (ODE) is defined by a time-dependent \emph{vector field} $u:\mathbb{R}^d\times[0,1]\to\mathbb{R}^d$ that specifies a velocity $u_t(x) \in \mathbb{R}^d$ for each position $x\in\mathbb{R}^d$ and time $t$. A solution to an ODE for a given initial condition $x_0$ is called a \emph{trajectory} $X:[0,1]\to\mathbb{R}^d$, describing the path of that single point over time with $X_t = x_t$. More generally, a \emph{flow} $\psi:\mathbb{R}^d\times[0,1]\to\mathbb{R}^d$ is the solution function of the ODE for all initial points $x_0$, satisfying
\[
\frac{d}{dt}\psi_t(x)=u_t(\psi_t(x)),\qquad \psi_0(x_0)=x_0.
\]

The goal of \emph{Flow Matching} (FM) is to learn a vector field $u_t^\theta$, parameterized by $\theta$, whose flow $\psi_t^\theta$ transports a simple base distribution $p_0 \triangleq p_{\mathrm{init}}$ to a target data distribution $p_1 \triangleq p_{\mathrm{data}}$. In practice, FM constructs a target velocity field $u_t^{\mathrm{target}}(x)$ from sample pairs and trains $u_t^\theta$ to match it, so that integrating $u_t^\theta(X_t)$ moves $p_{\mathrm{init}}$ along some predefined paths to generate samples from $p_{\mathrm{data}}$. These \emph{probability paths} $(p_t)_{0\leq t\leq1}$ specify a gradual interpolation between noise $p_{\mathrm{init}}$ and data $p_{\mathrm{data}}$. A standard probability path for example is the \emph{linear} path: given $X_0 \sim p_0$, $X_1 \sim p_1$ and $t\sim \mathrm{Unif}[0,1]$, we can sample $X_t = tX_1 + (1-t)X_0 \sim p_t$.\looseness-1

\noindent
The FM objective is
$$
    \mathcal{L}_{\mathrm{FM}}(\theta) = \mathbb{E}_{t\sim \mathrm{Unif}, x\sim p_t}[ ||u_t^\theta(x) - u_t^{target}(x)||^2], 
$$
but since the \emph{marginal} vector field $u_t^{\mathrm{target}}(x)$ is not tractable in practice, one minimizes the \emph{Conditional Flow Matching} (CFM) loss
$$
    \mathcal{L}_{\mathrm{CFM}}(\theta) = \mathbb{E}_{t\sim \mathrm{Unif}, z\sim p_{\mathrm{data}},x\sim p_t(\cdot|z)}[ ||u_t^\theta(x) - u_t^{target}(x|z)||^2], 
$$
because it holds that $\nabla \mathcal{L}_{\mathrm{FM}}=\nabla \mathcal{L}_{\mathrm{CFM}}$, where now $z$ is sampled from $p_{\mathrm{data}}$, $x$ is sampled from the \emph{conditional} probability path $p_t(\cdot|z)$, and $u_t^{target}(x|z)$ is the \emph{conditional} vector field. For a more complete explanation and formalization we remind~\cite{lipman2024flowmatchingguidecode}.

\subsection{Problem formulation}
\label{probform}
At time $\tau\!\ge\!0$ we represent the map as
$\mathcal M_\tau=(\mathcal O_\tau, O_{\mathrm{BG}})$, where
$\mathcal O_\tau=\{O^{i}_\tau\}_{i=1}^N$ is the set of objects, and $O_{\mathrm{BG}}$ denotes the static background that is assumed to remain unchanged over time. 
Each object is represented as
\[
O^{(i)}_\tau=\bigl(X^{i}_\tau, f^{i}_\tau, l^{i}\bigr),
\]

\noindent
where $X^i_\tau \in \mathbb{R}^2$ denotes its 2D position in space, $f^i_\tau$ represents object features (e.g., appearance or shape descriptors), and $l^i$ is a text descriptor (e.g., \texttt{red coffee mug}).

Given a prediction horizon $\Delta\tau\!\ge\!0$ and a text-object query $O^q$, our goal is to infer the likely (multimodal) future locations
at time $\tau_q=\tau+\Delta\tau$. Formally, we aim to compute the posterior
\begin{equation}
    \label{eq:objective}
p\!\left(X_{\tau_q}^{q}\,\middle|\,\mathcal M_\tau, O^q, \Delta\tau\right).
\end{equation}

As the posterior in \eqref{eq:objective} can be highly multimodal and intractable, we employ FM to approximate it and generate samples from it \cite{lipmanflow}. In particular, given a random sample $X_{\text{init}} \sim p_0$ obtained from a simple distribution $p_0 = \mathcal{N}(\mathbf{0}, \mathbf{I})$, our goal is to make $X_{\text{final}} \sim p_{\text{1}}$ with $p_1 = p\!\left(X_{\tau_q}^{q}\,\middle|\,\mathcal M_\tau, O^q, \Delta\tau\right)$.


\begin{figure}
  \centering
  \begin{subfigure}{0.23\textwidth}
    \includegraphics[width=\linewidth]{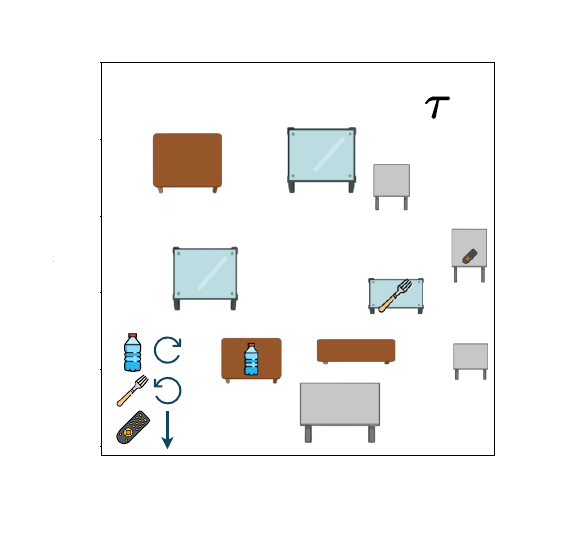}
  \end{subfigure}\hfill%
  \begin{subfigure}{0.24\textwidth}
    \includegraphics[width=\linewidth]{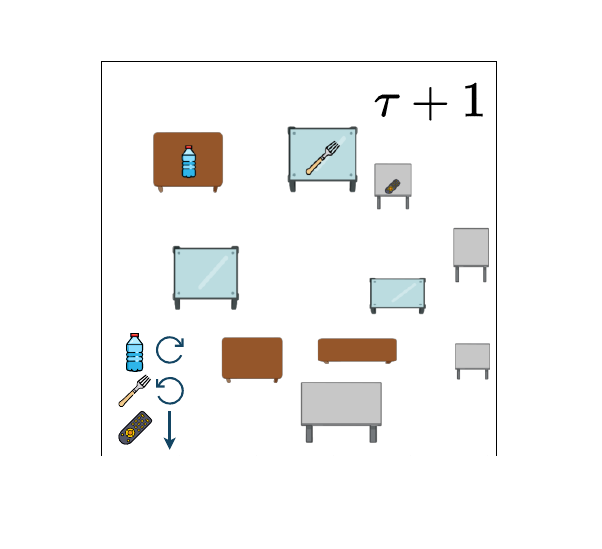}
  \end{subfigure}
  \vspace{-5mm}
  \caption{A FlowSim environment across two consecutive timestamps.}
  \vspace{-0.5cm}
  \label{fig:flowsim}
\end{figure}

\begin{figure*}[t!]
  \centering
    \begin{subfigure}[t]{0.33\textwidth}
    \centering
    \includegraphics[width=\linewidth]{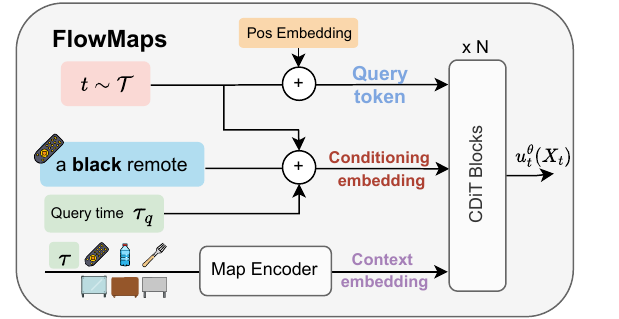}
    \caption{\name{}}
    \label{fig:fm}
  \end{subfigure}
  \begin{subfigure}[t]{0.31\textwidth}
    \centering
    \includegraphics[width=\linewidth]{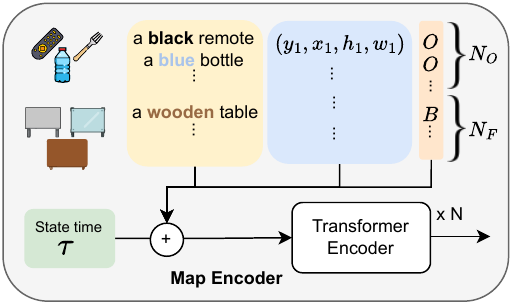}
    \caption{Map Encoder}
    \label{fig:map}
  \end{subfigure}\hspace{0.015\textwidth} 
  \begin{subfigure}[t]{0.29\textwidth}
    \centering
    \includegraphics[width=\linewidth]{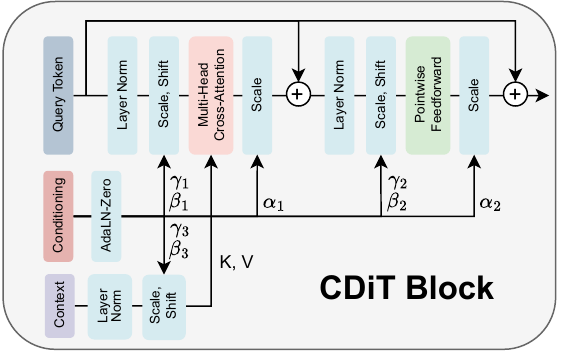}
    \caption{CDiT Block}
    \label{fig:cdit}
  \end{subfigure}
  \caption{\name{} (a) architecture and its modules, the (b) Map Encoder and the (c) CDiT Block.}
  \vspace{-0.5cm}
  \label{fig:arch}
\end{figure*}

\subsection{FlowSim}
\label{data}
FM (and generative models in general) are \emph{data-hungry}: they require a significant amount of data to be able to approximate well a target distribution. Despite recent community efforts toward large-scale dynamic datasets \cite{sun2025nothing}, to the best of our knowledge there is no resource that fully matches our needs: indoor scenes with objects that move along semantically consistent patterns over space and time.

To bridge this gap, we procedurally generate data with our proposed simulator, FlowSim. In FlowSim, objects move across static pieces of furniture following predefined, category-specific patterns. Figure~\ref{fig:flowsim} illustrates the motion between two consecutive timestamps, mimicking distinct human interaction patterns: a \emph{bottle} moving counterclockwise, a \emph{fork} clockwise, and a \emph{remote} top-to-bottom\footnote{We prototype the simulator with three object categories, but it is easily extensible to many more.}. Furthermore, transitions are stochastic: at each timestamp an object advances to the next location along its trajectory with probability \(0.7\), remains at where it is with probability \(0.2\), and skips one step (advancing by two locations) with probability \(0.1\). This controlled stochasticity induces multimodal spatio-temporal distributions.\looseness-1

FlowSim logs the bounding boxes and descriptors per timestamp for all objects in the scene (both static and dynamic) providing exactly the supervision required for \name{}. 
\subsection{\name{}}
\vspace{-0.3cm}
\label{arch}
In Fig.~\ref{fig:arch} the \name{} architecture is shown (\ref{fig:fm}), alongside its two main modules: the \textbf{Map Encoder} (\ref{fig:map}) and the \textbf{Conditional Diffusion Transformer} (CDiT) block (\ref{fig:cdit}). \looseness-1

The Map Encoder embeds the map at state time $\tau$ by encoding object and furniture descriptors (i.e., \emph{colors} or \emph{labels}) and bounding boxes into tokens $\mathbf{e}\in\mathbb{R}^{(N_O+N_F)\times D}$, where $N_O+N_F$ is the number of tokens. Similar to~\cite{wald2020learning}, we also encode an object-type flag indicating whether the $i$-th token is an object or a piece of furniture (denoted as $O$ and $B$ in Fig.~\ref{fig:map}). Tokens are padded to a maximum length $S$, augmented with a state time embedding of $\tau$, and passed through various transformer encoders~\cite{vaswani2017attention} to yield a scene-context map embedding. Although Fig.~\ref{fig:flowsim} shows a rendered environment, \name{} is \emph{not} a visual model: to avoid trivializing the task, it consumes FlowSim's structured annotations (bounding boxes and descriptors), not images.

The Diffusion Transformer (DiT) block~\cite{peebles2023scalable} is a scalable backbone for training latent generative models. We use a modified CDiT of \cite{bar2025navigation} with one change: we drop its Multi-Head Self-Attention, since we diffuse only the query object rather than the full scene. Each CDiT block takes (i) a \textcolor{query}{\textbf{query token}}, (ii) a \textcolor{cond}{\textbf{conditioning embedding}} that drives \emph{AdaLN-Zero} gates $(\alpha,\beta,\gamma)$, and (iii) a \textcolor{cont}{\textbf{context embedding}} used to cross-attend the \textcolor{query}{\textbf{query}}. To form a \textcolor{query}{\textbf{query token}}, we sample $t\sim U[0,1]$, interpolate $X_t$ along a probability path $p_t$ between $X_0\sim\mathcal{N}(\mathbf{0},\mathbf{I}_{4 \times 4})$ and the ground-truth box $X_1=(y,x,h,w)$, then add a sinusoidal positional embedding. The \textcolor{cond}{\textbf{conditioning embedding}} encodes and sums the query time $\tau_q$, the query object descriptor, and FM time $t$; the \textcolor{cont}{\textbf{context embedding}} is the Map Encoder output.\looseness-1

CDiT learns the vector field $u_t^{\theta}(X_t)$, and given ground-truth $u_t({X}_t)$, we minimize $\mathcal{L}=||u_t({X}_t)-u_t^\theta({X}_t)||^2$, backpropagating through the Map Encoder and the CDiT blocks.

During inference, we draw $X_0\sim\mathcal{N}(\mathbf{0},\mathbf{I}_{4\times4})$ and numerically integrate the ODE to obtain the predicted position $\hat{X}^{q}_{\tau_q}$ at future time $\tau_q$, conditioned on the current map $\mathcal{M}_{\tau}$ and the object descriptor.

\subsection{Implementation details}
\label{implementation}
We set embedding dimension to \(D=256\) and \(N=8\) blocks in both the CDiT and the Transformer encoder, maximum timestamp horizon $\tau_{max}=20$, and chose \emph{color} information as descriptors. We tested training with and without dropout and observed better performance with dropout enabled at \(p=0.1\). For the time-sampling distribution over \(t\), we evaluated the standard \emph{uniform}, \emph{beta}~\cite{black2024pi0}, and \emph{logit-normal}~\cite{esser2024scaling} distributions; only the \emph{uniform} choice degraded performance. As the probability path, we adopt the \emph{linear} path~\cite{lipman2024flowmatchingguidecode}. For ODE integration we use the \emph{midpoint} method~\cite{lipman2024flowmatchingguidecode}. We train with an \emph{AdamW} optimizer with $\beta_1 = 0.95$ and $\beta_2 = 0.999$, \emph{Exponential Moving Average} (EMA) smoothing for weights with rate of $0.9999$, and a \emph{cosine} learning-rate schedule with starting LR $\eta=1e-4$ for $30$k iterations. 

\section{RESULTS}
\begin{figure}[t!]
    \centering
    \includegraphics[width=0.9\linewidth]{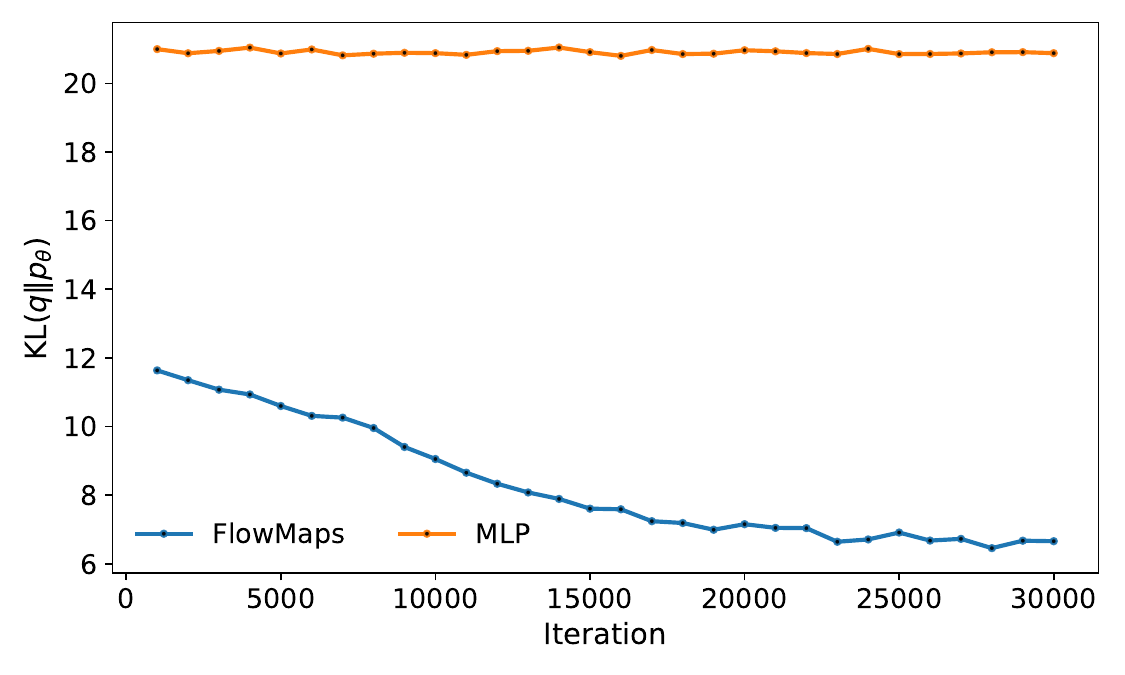}
    \vspace{-2mm}
    \caption{KL divergence comparison between FlowMaps (blue) and the MLP baseline (orange).}
    \label{fig:kl}
    \vspace{-3mm}
\end{figure}

\begin{figure}[t!]
  \centering
  \includegraphics[width=0.95\linewidth]{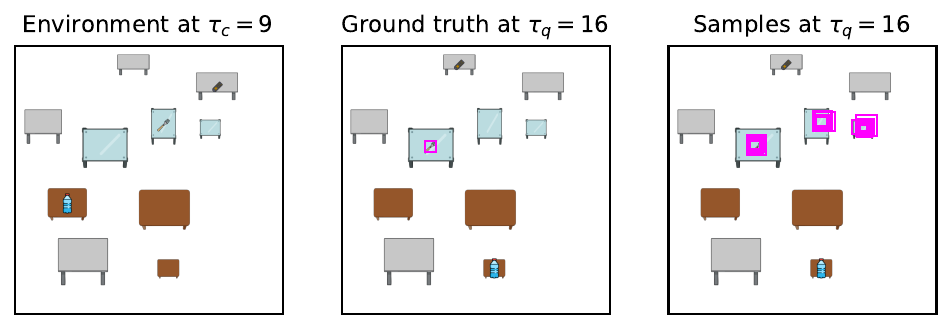}\\[4pt]
  \includegraphics[width=0.95\linewidth]{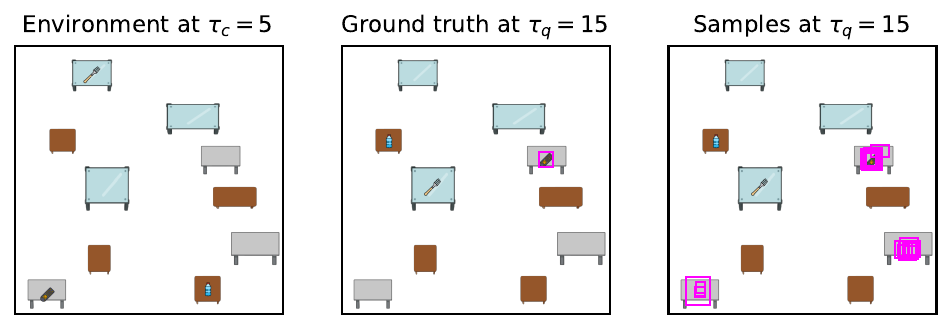}\\[4pt]
  \includegraphics[width=0.95\linewidth]{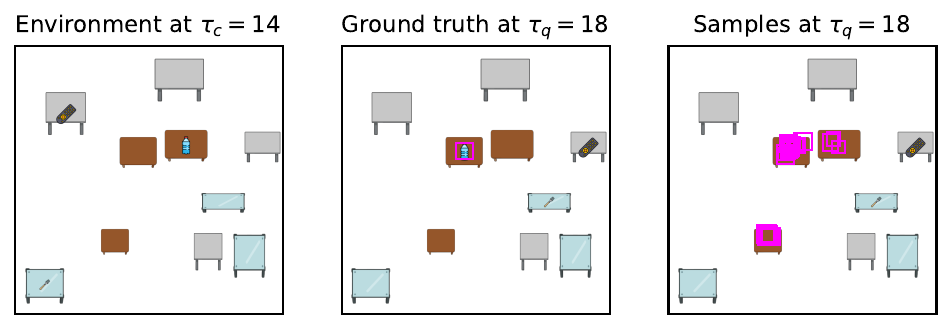}
  \vspace{-3mm}
  \caption{Results when predicting locations for a fork (top), remote (mid) and bottle (bottom).}
  \label{fig:res}
  \vspace{-0.5cm}
\end{figure}

We compare \name{} with an MLP-based baseline that also employs the same Map Encoder and regresses over a Gaussian distribution centered on the ground truth bounding box. Figure~\ref{fig:kl} reports the KL divergence $KL(q||p_\theta)$ computed between the ground truth distribution $q$ and the predicted one $p_\theta$. The results show that our method has better mode distribution coverage than the MLP baseline, highlighting the importance of a multimodal distributional approach for this task. Consistent with this trend, Fig.~\ref{fig:res} shows \name{} inference results for multiple objects at future timestamps. For each query, we generate 25 samples to obtain multiple plausible bounding box positions. As shown in the image, the model successfully distinguishes object-specific behaviors and recovers a multimodal distribution. Together, these qualitative and quantitative results indicate that \name{} effectively captures object-conditioned dynamics and the multimodal nature of future object positions.\looseness-1
\label{results}
\section{CONCLUSIONS}
\label{conclusion}
In this paper we introduced FlowMaps, a Flow Matching based model that infers multimodal posterior distributions over future object locations in household environments with long-term dynamics, together with FlowSim, a procedurally generated dataset that captures semantically consistent, human-induced object motions. 
Our results indicate that FlowMaps learns object-conditioned routines and provides robust priors that can help recover from missed relocalizations during object navigation and retrieval tasks. Looking ahead, we will (i) scale from simulation to real homes with longer horizons and richer object vocabularies and full text object queries, (ii) integrate FlowMaps as an uncertainty-aware placement prior within exploration and planning stacks, and (iii) extend the formulation to 3D Scene Graphs~\cite{gu2024conceptgraphs} and online updating so that the posterior can be refined as new evidence arrives. We believe this is a step toward robot behaviors that exploit human regularities to operate reliably in changing environments.




\section*{ACKNOWLEDGMENT}
This work has been carried out while Francesco Argenziano was enrolled in the Italian National Doctorate on Artificial Intelligence run by Sapienza University of Rome. This work was partially funded by an FRQNT B2X Scholarship with DOI 10.69777/335998 [MS]. This research was conducted while Francesco Argenziano was enrolled as a visiting researcher at Mila - Quebec AI Institute.
We acknowledge partial financial support from PNRR MUR project PE0000013-FAIR. 

\bibliographystyle{IEEEtran}
\bibliography{IEEEabrv.bib,  short-string.bib, references_short.bib}

\end{document}